%% file: acl2017.tex
\documentclass[11pt,a4paper]{article}
\usepackage[hyperref]{acl2017}
\usepackage{times}
\usepackage{latexsym}

\usepackage{amsmath}
\usepackage{multirow}
\usepackage{url}

\usepackage{booktabs}

\makeatother

\usepackage{graphicx}
\graphicspath{ {images/} }
\usepackage{caption}
\usepackage{subcaption}

\usepackage{amssymb}
\makeatletter
\renewcommand*{\p@section}{\S\,}
\renewcommand*{\p@subsection}{\S\,}

\usepackage{url}

\aclfinalcopy 
\def\aclpaperid{680} 

\title{Leveraging Knowledge Bases in LSTMs for Improving Machine Reading}
\author{Bishan Yang \\
  Machine Learning Department \\
  Carnegie Mellon University \\
  {\tt bishan@cs.cmu.edu} \\\And
  Tom Mitchell \\
  Machine Learning Department \\
  Carnegie Mellon University \\
  {\tt tom.mitchell@cs.cmu.edu} \\}

\date{}

\begin{document}
\maketitle
\begin{abstract}
This paper focuses on how to take advantage of external knowledge bases (KBs) to improve recurrent neural networks for machine reading. Traditional methods that exploit knowledge from KBs encode knowledge as discrete indicator features. Not only do these features generalize poorly, but they require task-specific feature engineering to achieve good performance. We propose KBLSTM, a novel neural model that leverages continuous representations of KBs to enhance the learning of recurrent neural networks for machine reading. To effectively integrate background knowledge with information from the currently processed text, our model employs an attention mechanism with a sentinel to adaptively decide whether to attend to background knowledge and which information from KBs is useful. Experimental results show that our model achieves accuracies that surpass the previous state-of-the-art results for both entity extraction and event extraction on the widely used ACE2005 dataset.
\end{abstract}

\input{sections/intro}
\input{sections/related_work}
\input{sections/method}
\input{sections/experiment}
\input{sections/conclusion}

\section*{Acknowledgments}
This research was supported in part by DARPA under contract number FA8750-13-2-0005, and by NSF grants IIS-1065251 and IIS-1247489. We also gratefully acknowledge the support of the Microsoft Azure for Research program and the AWS Cloud Credits for Research program. In addition, we would like to thank anonymous reviewers for their helpful comments.
\bibliography{acl2017}
\bibliographystyle{acl_natbib}

\end{document}

%% file: sections/intro.tex
\section{Introduction}
\label{sec:intro}
Recurrent neural networks (RNNs), a neural architecture that can operate over text sequentially, have shown great success in addressing a wide range of natural language processing problems, such as parsing~\cite{dyer:2015acl}, named entity recognition~\cite{lample2016neural}, and semantic role labeling~\cite{zhou2015end}). These neural networks are typically trained end-to-end where the input is only text or a sequence of words and a lot of background knowledge is disregarded.

The importance of background knowledge in natural language understanding has long been recognized~\cite{minsky1988society,fillmore1976frame}. Earlier NLP systems mostly exploited restricted linguistic knowledge such as manually-encoded morphological and syntactic patterns. With the advanced development of knowledge base construction, large amounts of semantic knowledge become available, ranging from manually annotated semantic networks like WordNet~\footnote{\url{https://wordnet.princeton.edu}} to semi-automatically or automatically constructed knowledge graphs like DBPedia~\footnote{\url{http://wiki.dbpedia.org/}} and NELL~\footnote{\url{http://rtw.ml.cmu.edu/rtw/kbbrowser/}}. While traditional approaches have exploited the use of these knowledge bases (KBs) in NLP tasks~\cite{ratinov2009design,rahman2011coreference,nakashole2015knowledge}, they require a lot of task-specific engineering to achieve good performance.

One way to leverage KBs in recurrent neural networks is by augmenting the dense representations of the networks with the symbolic features derived from KBs. This is not ideal as the symbolic features have poor generalization ability. In addition, they can be highly sparse, e.g., using WordNet synsets can easily produce millions of indicator features, leading to high computational cost. Furthermore, the usefulness of knowledge features varies across contexts, as general KBs involve polysemy, e.g., ``Clinton" can refer to a person or a town. Incorporating KBs irrespective of the textual context could mislead the machine reading process.

Can we train a recurrent neural network that learns to adaptively leverage knowledge from KBs to improve machine reading? In this paper, we propose KBLSTM, an extension to bidirectional Long Short-Term Memory neural networks (BiLSTMs)~\cite{hochreiter1997long,graves2005bidirectional} that is capable of leveraging symbolic knowledge from KBs as it processes each word in the text. At each time step, the model retrieves KB concepts that are potentially related to the current word. Then, an attention mechanism is employed to dynamically model their semantic relevance to the reading context. Furthermore, we introduce a sentinel component in BiLSTMs that allows flexibility in deciding whether to attend to background knowledge or not. This is crucial because in some cases the text context should override the context-independent background knowledge available in general KBs.

In this work, we leverage two general, readily available knowledge bases: WordNet~\cite{WordNet} and NELL~\cite{NELL-aaai15}. WordNet is a manually created lexical database that organizes a large number of English words into sets of synonyms (i.e. synsets) and records conceptual relations (e.g., hypernym, part\_of) among them. NELL is an automatically constructed web-based knowledge base that stores beliefs about entities. It is organized based on an ontology of hundreds of semantic categories (e.g., person, fruit, sport) and relations (e.g., personPlaysInstrument). We learn distributed representations (i.e., embeddings) of WordNet and NELL concepts using knowledge graph embedding methods. We then integrate these learned embeddings with the state vectors of the BiLSTM network to enable knowledge-aware predictions.

We evaluate the proposed model on two core information extraction tasks: entity extraction and event extraction. For entity extraction, the model needs to recognize all mentions of entities such as person, organization, location, and other things from text. For event extraction, the model is required to identify event mentions or event triggers\footnote{An event also consists of participants whose types depend on the event triggers. In this work, we focus on predicting event triggers and leave the prediction of event participants for future work.} that express certain types of events, e.g., elections, attacks, and travels. Both tasks are challenging and often require the combination of background knowledge and the text context for accurate prediction. For example, in the sentence ``Maigret left viewers in tears.", knowing that ``Maigret" can refer to a TV show can greatly help disambiguate its meaning. However, knowledge bases may hurt performance if used blindly. For example, in the sentence ``Santiago is charged with murder.", methods that rely heavily on KBs are likely to interpret ``Santiago" as a location due to the popular use of Santiago as a city. Similarly for events, the same word can trigger different types of events, for example, ``release" can be used to describe different events ranging from book publishing to parole. It is important for machine learning models to learn to decide which knowledge from KBs is relevant given the context. 

Extensive experiments demonstrate that our KBLSTM models effectively leverage background knowledge from KBs in training BiLSTM networks for machine reading. They achieve significant improvement on both entity and event extraction compared to traditional feature-based methods and LSTM networks that disregard knowledge in KBs, resulting in new state-of-the-art results for entity extraction and event extraction on the widely used ACE2005 dataset.

%% file: sections/related_work.tex
\section{Related Work}
Essential to RNNs' success on natural language processing is the use of Long Short-Term Memory neural networks~\cite{hochreiter1997long} (LSTMs) or Gated Recurrent Unit~\cite{cho2014learning} (GRU) as they are able to handle long-term dependencies by adaptively memorizing values for either long or short durations. Their bidirectional variants BiLSTM~\cite{graves2005bidirectional} or BiGRU further allow the incorporation of both past and future information. Such ability has been shown to be generally helpful in various NLP tasks such as named entity recognition~\cite{huang2015bidirectional,chiu2016named,ma2016end}, semantic role labeling~\cite{zhou2015end}, and reading comprehension~\cite{hermann2015teaching,chen2016thorough}. In this work, we also employ the BiLSTM architecture.

In parallel to the development for text processing, neural networks have been successfully used to learn distributed representations of structured knowledge from large KBs~\cite{bordes2011learning,bordes2013translating,socher2013reasoning,yang2014embedding,guu2015traversing}. Embedding the symbolic representations into continuous space not only makes KBs more easy to use in statistical learning approaches, but also offers strong generalization ability. Many attempts have been made on connecting distributed representations of KBs with text in the context of knowledge base completion~\cite{lao2011random,gardner2014incorporating,toutanova2015representing}, relation extraction~\cite{weston2013connecting,chang2014typed,riedel2013relation}, and question answering~\cite{miller2016key}. However, these approaches model text using shallow representations such as subject/relation/object triples or bag of words. More recently,~\newcite{ahn2016neural} proposed a neural knowledge language model that leverages knowledge bases in RNN language models, which allows for better representations of words for language modeling. Unlike their work, we leverage knowledge bases in LSTMs and applies it to information extraction.

The architecture of our KBLSTM model draws on the development of attention mechanisms that are widely employed in tasks such as machine translation~\cite{bahdanau2014neural} and image captioning~\cite{xu2015show}. Attention allows the neural networks to dynamically attend to salient features of the input. With a similar motivation, we employ attention in KBLSTMs to allow for dynamic attention to the relevant knowledge given the text context. Our model is also closely related to a recent model of caption generation introduced by~\newcite{lu2016knowing}, where a visual sentinel is introduced to allow the decoder to decide whether to attend to image information when generating the next word. In our model, we introduce a sentinel to control the tradeoff between background knowledge and information from the text. 


%% file: sections/method.tex
\section{Method}
\label{sec:method}
In this section, we present our KBLSTM model. We first describe several basic recurrent neural network frameworks for machine reading, including basic RNNs, LSTMs, and bidirectional LSTMs (Sec.~\ref{sec:rnn}).  We then introduce our extension to bidirectional LSTMs that allows for the incorporation of KB information at each time step of reading (Sec.~\ref{sec:KBLSTM}). The KB information is encoded using continuous representations (i.e., embeddings) which are learned using knowledge embedding methods (Sec.~\ref{sec:kb-emb}).

\subsection{RNNs, LSTMs, and Bidirectional LSTMs}
\label{sec:rnn}
RNNs are a class of neural networks that take a sequence of inputs and compute a hidden state vector at each time step based on the current input and the entire history of inputs. The hidden state vector can be computed recursively using the following equation~\cite{elman1990finding}: 
$${\bf h}_t=F({\bf W}{\bf h}_{t-1}+{\bf U}{\bf x}_t)$$
where ${\bf x}_t$ is the input at time step $t$, ${\bf h}_t$ is the hidden state at time step $t$, ${\bf U}$ and ${\bf W}$ are weight matrices, and $F$ is a nonlinear function such as tanh or ReLu. Depending on the applications, RNNs can produce outputs based on the hidden state of each time step or just the last time step. 

A Long Short-Term Memory network~\cite{hochreiter1997long} (LSTM) is a variant of RNNs which was design to better handle cases where the output at time $t$ depends on much earlier inputs. It has a memory cell and three gating units: an input gate that controls what information to add to the current memory, a forget gate which controls what information to remove from the previous memory, and an output gate which controls what information to output from the current memory. Each gate is implemented as a logistic function $\sigma$ that takes as input the previous hidden state and the current input, and outputs values between 0 and 1. Multiplication with these values controls the flow of information into or out of the memory. In equations, the updates at each time step $t$ are:
\begin{equation*}
\begin{split}
& {\bf i}_t = \sigma({\bf W}_i{\bf h}_{t-1} + {\bf U}_i{\bf x}_t) \\
& {\bf f}_t = \sigma({\bf W}_f{\bf h}_{t-1} + {\bf U}_f{\bf x}_t) \\
& {\bf o}_t = \sigma({\bf W}_o{\bf h}_{t-1} + {\bf U}_o{\bf x}_t) \\
& {\bf c}_t = {\bf f}_t\odot {\bf c}_{t-1} + {\bf i}_t\odot tanh({\bf W}_c{\bf h}_{t-1} + {\bf U}_c{\bf x}_t) \\
& {\bf h}_t = {\bf o}_t\odot tanh({\bf c}_t)\\
\end{split}
\end{equation*}
where ${\bf i}_t$ is the input gate, ${\bf f}_t$ is the forget gate, ${\bf o}_t$ is the output gate, ${\bf c}_t$ is the memory cell, and ${\bf h}_t$ is the hidden state. $\odot$ denotes element-wise multiplication. ${\bf W}_i,{\bf U}_i,{\bf W}_f,{\bf U}_f,{\bf W}_o,{\bf U}_o,{\bf W}_c,{\bf U}_c$ are weight matrices to be learned. 

Bidirectional LSTMs~\cite{graves2005bidirectional} (BiLSTMs) are essentially a combination of two LSTMs in two directions: one operates in the forward direction and the other operates in the backward direction. This leads to two hidden states $\overrightarrow{{\bf h}_t}$ and $\overleftarrow{{\bf h}_t}$ at time step $t$, which can be viewed as a summary of the past and the future respectively. Their concatenation ${\bf h}_t=[\overrightarrow{{\bf h}_t};\overleftarrow{{\bf h}_t}]$ provides a whole summary of the information about the input around time step $t$. Such property is attractive in NLP tasks, since information of both previous words and future words can be helpful for interpreting the meaning of the current word.     

\subsection{Knowledge-aware Bidirectional LSTMs}
\label{sec:KBLSTM}
\begin{figure}
\centering
\includegraphics[width=0.5\textwidth]{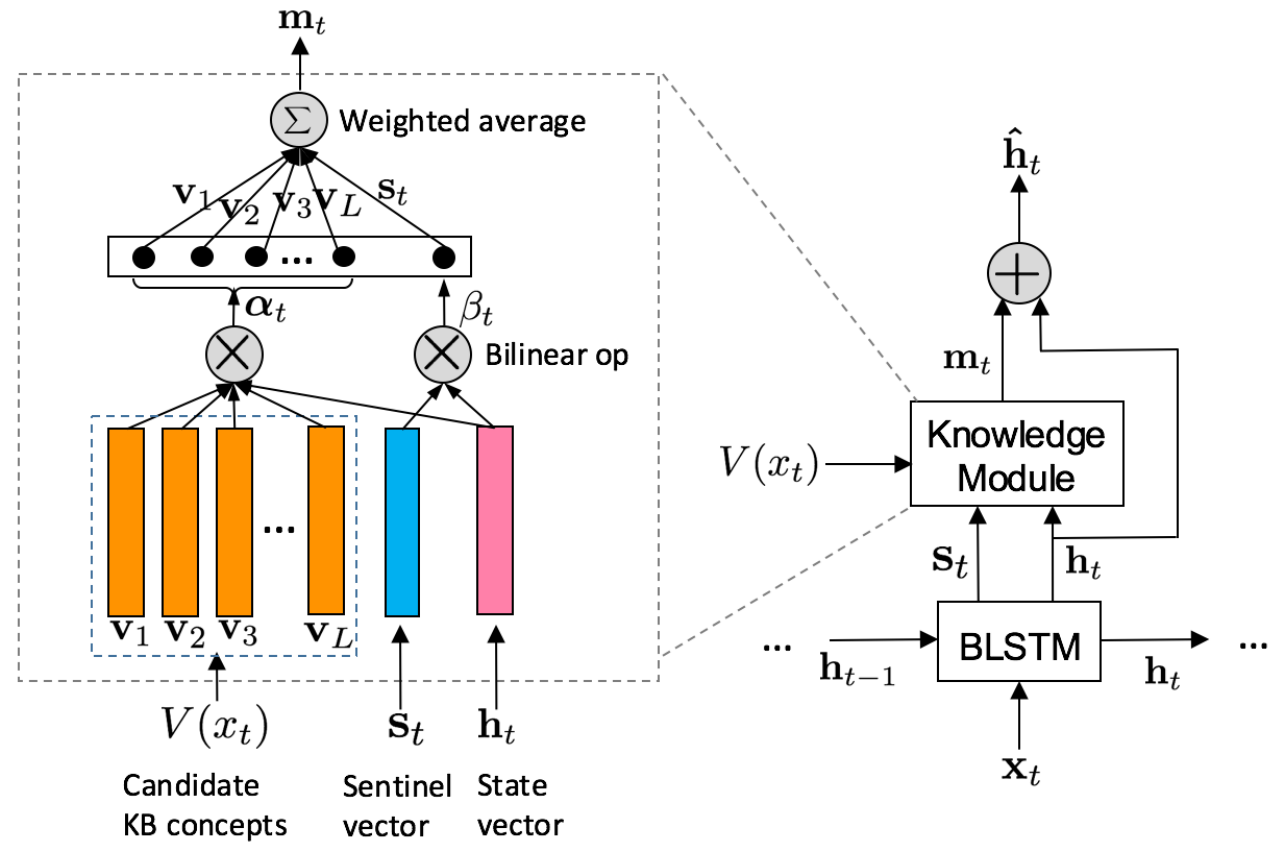}
\caption{{Architecture of the KBLSTM model. As each time step $t$, the knowledge module retrieves a set of candidate KB concepts $V(x_t)$ that are related to the current input $x_t$, and then computes a knowledge state vector ${\bf m}_t$ that integrates the embeddings of the candidate KB concepts ${\bf v}_1,{\bf v}_2,...,{\bf v}_L$ and the current context vector ${\bf s}_t$. See Section~\ref{sec:KBLSTM} for details.} }
\label{fig:KBLSTM}
\end{figure}

Our model (referred to as KBLSTM) extends BiLSTMs to allow flexibility in incorporating symbolic knowledge from KBs. Instead of encoding knowledge as discrete features, we encode it using continuous representations. Concretely, we learn embeddings of concepts in KBs using a knowledge graph embedding method. (We will describe the details in Section~\ref{sec:kb-emb}). The KBLSTM model then retrieves the embeddings of candidate concepts that are related to the current word being processed and integrates them into its state vector to make knowledge-aware predictions. Figure~\ref{fig:KBLSTM} depicts the architecture of our model. 

The core of our model is the knowledge module, which is responsible for transferring background knowledge into the BiLSTMs. The knowledge at time step $t$ consists of candidate KB concepts $V(x_t)$ for input $x_t$. (We will describe how to obtain the candidate KB concepts from NELL and WordNet in Section~\ref{sec:kb-emb}). Each candidate KB concept $i\in V(x_t)$ is associated with a vector embedding ${\bf v}_i$. We compute an attention weight $\alpha_{ti}$ for concept vector ${\bf v}_i$ via a bilinear operator, which reflects how relevant or important concept $i$ is to the current reading context ${\bf h}_t$:
\begin{equation}
\alpha_{ti}\propto\exp({\bf v}_i^T{\bf W}_v{\bf h}_t)
\end{equation}
where ${\bf W}_v$ is a parameter matrix to be learned. 

Note that the candidate concepts in some cases are misleading. For example, a KB may store the fact that ``Santiago" is a city but miss the fact that it can also refer to a person. Incorporating such knowledge in the sentence ``Santiago is charged with murder." could be misleading. To address this issue, we introduce a knowledge sentinel that records the information of the current context and use a mixture model to allow for better tradeoff between the impact of background knowledge and information from the context. Specifically, we compute a sentinel vector ${\bf s}_t$ as:
\begin{equation}
{\bf b}_t=\sigma({\bf W}_b{\bf h}_{t-1} + {\bf U}_b{\bf x}_t)
\end{equation}
\begin{equation}
{\bf s}_t={\bf b}_t\odot tanh({\bf c}_t)
\end{equation}
where ${\bf W}_b$ and ${\bf U}_b$ are weight parameters to be learned. The weight on the local context is computed as:
\begin{equation}
\beta_t\propto \exp({\bf s}_t^T{\bf W}_s{\bf h}_t)
\end{equation}
where ${\bf W}_s$ is a parameter matrix to be learned. The mixture model is defined as: 
\begin{equation}
{\bf m}_t=\sum_{i\in V(x_t)}\alpha_{ti}{\bf v}_i + \beta_t{\bf s}_t
\end{equation}
where $\sum_{i\in V(x_t)}\alpha_{ti}+\beta_t=1$. ${\bf m}_t$ can be viewed as a knowledge state vector that encodes external KB information with respect to the input at time $t$. We combine it with the state vector ${\bf h}_t$ of BiLSTMs to obtain a knowledge-aware state vector ${\bf \hat{h}}_t$:
\begin{equation}
\label{eq:kbvec}
{\bf \hat{h}}_t = {\bf h}_t + {\bf m}_t
\end{equation}
If $V(x_t)=\emptyset$, we set ${\bf m}_t=0$. ${\bf \hat{h}}_t$ can be used for predictions in the same way as the original state vector ${\bf h}_t$ (see Section~\ref{sec:exp} for details). 

\subsection{Embedding Knowledge Base Concepts}
\label{sec:kb-emb}
Now we describe how to learn embeddings of concepts in KBs. We consider two KBs: WordNet and NELL, which are both knowledge graphs that can be stored in the form of RDF\footnote{\url{https://www.w3.org/TR/rdf11-concepts/}} triples. Each triple consists of a subject entity, a relation, and an object entity. Examples of triples in WordNet are (\textit{location}, \textit{hypernym\_of}, \textit{city}), and (\textit{door}, \textit{has\_part}, \textit{lock}), where both the subject and object entities are synsets in WordNet. Examples of triples in NELL are (\textit{New York}, \textit{located\_in}, \textit{United States}) and (\textit{New York}, \textit{is\_a}, \textit{city}), where the subject entity is a noun phrase that can refer to a real-world entity and the object entity can be either a noun phrase entity or a concept category. 

In this work, we refer to the synsets in WordNet and the concept categories in NELL as \textit{KB concepts}. They are the key components of the ontologies and provide generally useful information for language understanding. As our KBLSTM model reads through each word in a sentence, it retrieves knowledge from NELL by searching for entities with the current word and collecting the related concept categories as candidate concepts; and it retrieves knowledge from WordNet by treating the synsets of the current word as candidate concepts.

We employ a knowledge graph embedding approach to learn the representations of the candidate concepts. Denote a KB triple as $(e_1,r,e_2)$, we want to learn embeddings of the subject entity $e_1$, the object entity $e_2$, and the relation $r$, so that the relevance of the triple can be measured by a scoring function based on the embeddings. We employ the {\sc Bilinear} model described in~\cite{yang2014embedding}.\footnote{We also experimented with TransE~\cite{bordes2013translating} and NTN~\cite{socher2013reasoning}, and found that they perform significantly worse than the Bilinear method, especially on predicting the ``is\_a" facts in NELL.} It computes the score of a triple $(e_1,r,e_2)$ via a bilinear function: $S_{(e_1, r, e_2)}={\bf v}_{e_1}^T{\bf M}_r{\bf v}_{e_2}$, where ${\bf v}_{e_1}$ and ${\bf v}_{e_2}$ are vector embeddings for $e_1$ and $e_2$ respectively, and ${\bf M}_r$ is a relation-specific embedding matrix. We train the embeddings using the max-margin ranking objective:
\begin{equation}
\label{obj:rank}
\sum_{q=(e_1,r,e_2)\in\mathcal{T}}\sum_{q'=(e_1,r,e_2')\in \mathcal{T}'}\max\{0,1-S_{q}+S_{q'}\}
\end{equation}
where $\mathcal{T}$ denotes the set of triples in the KB and $\mathcal{T}'$ denotes the ``negative" triples that are not observed in the KB. 

For WordNet, we train the concept embeddings using the preprocessed data provided by~\cite{bordes2013translating}, which contains 151,442 triples with 40,943 synsets and 18 relations. We use the same data splits for training, development, and testing. During training, we use AdaGrad~\cite{duchi2011adaptive} to optimize objective~\ref{obj:rank} with an initial learning rate of 0.05 and a mini-batch size of 100. At each gradient step, we sample 10 negative object entities with respect to the positive triple. Our implementation of the {\sc Bilinear} model achieves top-10 accuracy of 91\% for predicting missing object entities on the WordNet test set, which is comparable with previous work~\cite{yang2014embedding}. 

For NELL, we train the concept embeddings using a subset of the NELL data\footnote{\url{http://rtw.ml.cmu.edu/rtw/resources}}. We filter noun phrases with annotation confidence less than 0.9 in order to remove erroneous labels introduced during the automatic construction of NELL~\cite{wijaya2016verbkb}. This results in 180,107 noun phrases and 258 concept categories in total. We randomly split 80\% of the data for training, 10\% for development and 10\% for testing. For each training example, we enumerate all the unobserved concept categories as negative labels. Instead of treating each entity as a unit, we represent it as an average of its constituting word vectors concatenated with its head word vector. The word vectors are initialized with pre-trained paraphrastic embeddings~\cite{wieting2015towards} and fine-tuned during training. Using the same optimization parameters as before, the {\sc Bilinear} model achieves 88\% top-1 accuracy for predicting concept categories of given noun phrases on the test set.

%% file: sections/experiment.tex
\section{Experiments}
\label{sec:exp}
\subsection{Entity Extraction}
\label{sec:entity_exp}
We first apply our model to entity extraction, a task that is typically addressed by assigning each word/token BIO labels ($Begin$, $Inside$, and $Outside$)~\cite{ratinov2009design} indicating the token's position within an entity mention as well as its entity type.

To allow tagging over phrases instead of words, we address entity extraction in two steps. The first step detects mention chunks, and the second step assigns entity type labels to mention chunks (including an {\sc O} type indicating \textit{other} types). In the first stage, we train a BiLSTM network with a conditional random field objective~\cite{huang2015bidirectional} using gold-standard BIO labels regardless of entity types, and only predict each token's position within an entity mention. This produces a sequence of chunks for each sentence. In the second stage, we train another supervised BiLSTM model to predict type labels for the previously extracted chunks. Each chunk is treated as a unit input to the BiLSTMs and the input vector is computed by averaging the input vectors of individual words within the chunk concatenated with its head word vector. The BiLSTMs can be trained with a softmax objective that minimizes the cross-entropy loss for each individual chunk. It computes the probability of the correct type for each chunk:
\begin{equation}
\label{obj:softmax}
p_{y_t}=\frac{\exp({\bf w}_{y_t}^T {\mathbf{h}_t})}{\sum_{y'_t}\exp({\bf w}_{y'_t}^T {\mathbf{h}_t})}
\end{equation}
The BiLSTMs can also be trained with a CRF objective (referred to as BiLSTM-CRF) that minimizes the negative log-likelihood for the entire sequence. It computes the probability of the correct types for a sequence of chunks:
\begin{equation}
\label{obj:crf}
p_{\bf y}=\frac{\exp(g({\bf x},{\bf y}))}{\sum_{{\bf y'}}\exp(g({\bf x}, {\bf y'}))}
\end{equation}
where $g({\bf x}, {\bf y})=\sum_{t=1}^l P_{t,y_t}+\sum_{t=0}^l A_{y_t,y_{t+1}}$, $P_{t,y_t}={\bf w}_{y_t}^T {\mathbf{h}_t}$ is the score of assigning the $t$-th chunk with tag $y_t$ and $A_{y_t, y_{t+1}}$ is the score of transitioning from tag $y_t$ to $y_{t+1}$. By replacing ${\bf h}_t$ in Eq.~\ref{obj:softmax} and Eq.~\ref{obj:crf} with the knowledge-aware state vector ${\bf \hat{h}}_t$ (Eq.~\ref{eq:kbvec}), we can compute the objective for KBLSTM and KBLSTM-CRF respectively. 
\subsubsection{Implementation Details}
\label{sec:expset}
We evaluate our models on the ACE2005 corpus~\cite{ACE2005} and the OntoNotes 5.0 corpus~\cite{hovy2006ontonotes} for entity extraction. Both datasets consist of text from a variety of sources such as newswire, broadcast conversations, and web text. We use the same data splits and task settings for ACE2005 as in~\newcite{li2014constructing} and for OntoNotes 5.0 as in~\newcite{durrett2014joint}.

At each time step, our models take as input a word vector and a capitalization feature~\cite{chiu2016named}. We initialize the word vectors using pretrained paraphrastic embeddings~\cite{wieting2015towards}, as we find that they significantly outperforms randomly initialized embeddings. The word embeddings are fine-tuned during training. For the KBLSTM models, we obtain the embeddings of KB concepts from NELL and WordNet as described in Section~\ref{sec:kb-emb}. These embeddings are kept fix during training. 

We implement all the models using Theano on a single GPU. We update the model parameters on every training example using Adam with default settings~\cite{kingma2014adam} and apply dropout to the input layer of the BiLSTM with a rate of 0.5. The word embedding dimension is set to 300 and the hidden vector dimension is set to 100. We train models on ACE2005 for about 5 epochs and on OntoNotes 5.0 for about 10 epochs with early stopping based on development results.

For each experiment, we report the average results over 10 random runs. We also apply the Wilcoxon rank sum test to compare our models with the baseline models. 

\subsubsection{Results}
We compare our KBLSTM and KBLSTM-CRF models with the following baselines: BiLSTM, a vanilla BiLSTM network trained using the same input, and BiLSTM-Fea, a BiLSTM network that combines its hidden state vector with discrete KB features (i.e., indicators of candidate KB concepts) to produce the final state vector. We also include their variants BiLSTM-CRF and BiLSTM-Fea-CRF, which are trained using the CRF objective instead of the standard softmax objective.

We first report results on entity extraction with gold-standard boundaries for multi-word mentions. This allows us to focus on the performance of entity type prediction without considering mention boundary errors and the noise they introduce in retrieving candidate KB concepts. Table~\ref{table:entity_results} shows the results.\footnote{$*$ indicates $p<0.05$ when comparing to the BiLSTM-based models.} We find that the CRF objective generally outperforms the softmax objective. Our KBLSTM-CRF model significantly improves over its counterpart BiLSTM-Fea-CRF. This demonstrates the effectiveness of KBLSTM architecture in leveraging KB information. 

Table~\ref{table:entity_results_kbs} breaks down the results of the KBLSTM-CRF and the BiLSTM-Fea-CRF using different KB settings. We find that the KBLSTM-CRF outperforms the BiLSTM-Fea-CRF in all the settings and that incorporating both KBs leads to the best performance. 

\begin{table}
\begin{footnotesize}
\begin{center}
\begin{tabular}{lccc}
\toprule
{\bf Model} & {\bf P} & {\bf R} & {\bf F1} \\
\midrule
BiLSTM & 83.5 & 86.4 & 84.9 \\

BiLSTM-CRF & 87.3 & 84.7 & 86.0 \\

BiLSTM-Fea & 86.1 & 84.7 & 85.4\\

BiLSTM-Fea-CRF & 87.7 & 86.1 & 86.9\\
\midrule
KBLSTM & 87.8 & 86.6 & 87.2 \\

KBLSTM-CRF & {\bf 88.1} & {\bf 87.8} & {\bf 88.0}$^*$ \\
\bottomrule
\end{tabular}
\end{center}
\caption{\label{table:entity_results} Entity extraction results on the ACE2005 test set with gold-standard mention boundaries.}
\end{footnotesize}
\end{table}

\begin{table}
\begin{footnotesize}
\begin{center}
\begin{tabular}{lcccc}
\toprule
{\bf Model} & {\bf KB} & {\bf P} & {\bf R} & {\bf F1} \\
\midrule
\multirow{3}{*}{BiLSTM-Fea-CRF} & NELL & 87.2 & 86.1 &86.6 \\
& WordNet & 86.4 & 86.0 & 86.2\\
& Both & 87.7 & 86.1 & 86.9\\
\midrule
\multirow{3}{*}{KBLSTM-CRF} & NELL & 87.4 & 87.6 & 87.5\\
& WordNet & 87.1 & 87.4 & 87.3\\
& Both & {\bf 88.1} & {\bf 87.8} & {\bf 88.0}\\
\bottomrule
\end{tabular}
\end{center}
\caption{\label{table:entity_results_kbs} Ablation results with different KBs.}
\end{footnotesize}
\end{table}

Next, we evaluate our models on entity extraction with predicted mention boundaries. We first train a BiLSTM-CRF to perform mention chunking using the same setting as described in Section~\ref{sec:expset}. We then treat the predicted chunks as units for entity type labeling. Table~\ref{table:ace_entity} reports the full entity extraction results on the ACE2005 test set. We compare our models with the state-of-the-art feature-based linear models~\newcite{li2014constructing},~\newcite{yang2016joint}, and the recently proposed sequence- and tree-structured LSTMs~\cite{miwa2016end}. Interestingly, we find that using BiLSTM-CRF without any KB information already gives strong performance compared to previous work. The KBLSTM-CRF model demonstrates the best performance among all the models and achieves the new state-of-the-art performance on the ACE2005 dataset.

We also report the entity extraction results on the OntoNotes 5.0 test set in Table~\ref{table:onto_entity}. We compare our models with the existing feature-based models~\newcite{ratinov2009design} and~\newcite{durrett2014joint}, which both employ heavy feature engineering to bring in external knowledge. BiLSTM-CNN~\cite{chiu2016named} employs a hybrid BiLSTM and Convolutional neural network (CNN) architecture and incorporates rich lexicon features derived from SENNA and DBPedia. Our KBLSTM-CRF model shows competitive results compared to their results. It also presents significant improvements compared to the BiLSTM and BiLSTM-Fea models. Note that the benefit of adding KB information is smaller on OntoNotes compared to ACE2005. We believe that there are two main reasons. One is that NELL has a lower coverage of entity mentions in OntoNotes than in ACE2005 (57\% vs. 65\%). Second, OntoNotes has a significantly larger amount of training data, which allows for more accurate models without much help from external resources. 

\begin{table}
\begin{footnotesize}
\begin{center}
\begin{tabular}{lccc}
\toprule
Model & P & R & F1\\
\midrule
\newcite{li2014incremental} & 85.2 & 76.9 & 80.8\\
\newcite{yang2016joint} & 83.5 & 80.2 & 81.8\\
\newcite{miwa2016end} & 82.9 & 83.9 & 83.4\\
\midrule
BiLSTM & 82.5 & 83.1 & 82.8\\
BiLSTM-CRF & 84.6 & 82.5 & 83.6\\
BiLSTM-Fea & 84.3 & 83.2 & 83.7\\
BiLSTM-Fea-CRF &84.7 &83.5  &84.1\\
KBLSTM & 85.5 & 85.2 & 85.3\\
KBLSTM-CRF & {\bf 85.4} & {\bf 86.0} & {\bf 85.7}$^*$ \\
\bottomrule
\end{tabular}
\end{center}
\caption{\label{table:ace_entity}Entity extraction results on the ACE2005 test set.}
\end{footnotesize}
\end{table}

\begin{table}
\begin{footnotesize}
\begin{center}
\begin{tabular}{lccc}
\toprule
Model & P & R & F1\\
\midrule
\newcite{ratinov2009design} & 82.0 & 84.9 & 83.4\\
\newcite{durrett2014joint} & 85.2 & 82.8 & 84.0\\
BiLSTM-CNN & 82.5 & 82.4 & 82.5\\
BiLSTM-CNN+emb & 85.9 & 86.3 & 86.1\\
BiLSTM-CNN+emb+lexicon & 86.0 & 86.5 & 86.2\\
\midrule
BiLSTM & 84.9 & 86.3 & 85.6\\
BiLSTM-CRF & 85.3 & 86.6 & 85.9\\
BiLSTM-Fea & 85.2 & 86.4 & 85.8\\
BiLSTM-Fea-CRF & 85.2 & 86.8 & 86.0\\
KBLSTM & {\bf 86.3} & 86.2 & 86.2  \\
KBLSTM-CRF & 86.1 & {\bf 86.8} & {\bf 86.4}$^*$\\
\bottomrule
\end{tabular}
\end{center}
\caption{\label{table:onto_entity}Entity extraction results on the OntoNotes 5.0 test set.}
\end{footnotesize}
\end{table}

\subsection{Event Extraction}
We now apply our model to the task of event extraction. Event extraction is concerned with detecting \textit{event triggers}, i.e., a word that expresses the occurrence of a pre-defined type of event. Event triggers are mostly verbs and eventive nouns but can occasionally be adjectives and other content words. Therefore, the task is typically addressed as a classification problem where the goal is to label each word in a sentence with an event type or an {\sc O} type if it does not express any of the defined events. It is straightforward to apply the BiLSTM architecture to event extraction. Similarly to the models for entity extraction, we can train the BiLSTM network with both the softmax objective and the CRF objective. 

We evaluate our models on the portion ACE2005 corpus that has event annotations. We use the same data split and experimental setting as in~\newcite{li2013joint}. The training procedure is the same as in Section~\ref{sec:expset}, and we train all the models for about 5 epochs. For the KBLSTM models, we integrate the learned embeddings of WordNet synsets during training.

\begin{table}
\begin{footnotesize}
\begin{center}
\begin{tabular}{lccc}
\toprule
Model & P & R & F1\\
\midrule
{\sc JointBeam} & 74.0 & 56.7 & 64.2\\
{\sc JointEventEntity} & {\bf 75.1} & 63.3 & 68.7\\
\midrule
{\sc DMCNN} & 71.8 & 63.8 & 69.0\\
{\sc JRNN} & 66.0 & 73.0 & 69.3\\
\midrule
BiLSTM & 71.3 & 59.3 & 64.7\\
BiLSTM-CRF & 64.2 & 66.6 & 65.4\\
BiLSTM-Fea & 68.4 & 62.7 & 65.5\\
BiLSTM-Fea-CRF & 65.5 & 66.7 & 66.1\\
KBLSTM & 70.1 & 67.3 & 68.7\\
KBLSTM-CRF & 71.6 & 67.8 & {\bf 69.7}$^*$ \\
\bottomrule
\end{tabular}
\end{center}
\caption{\label{table:event_results}event extraction results on the ACE2005 test set.}
\end{footnotesize}
\end{table}

\begin{figure*}  
\centering
        \begin{subfigure}{.45\textwidth}
             \includegraphics[width=\linewidth]{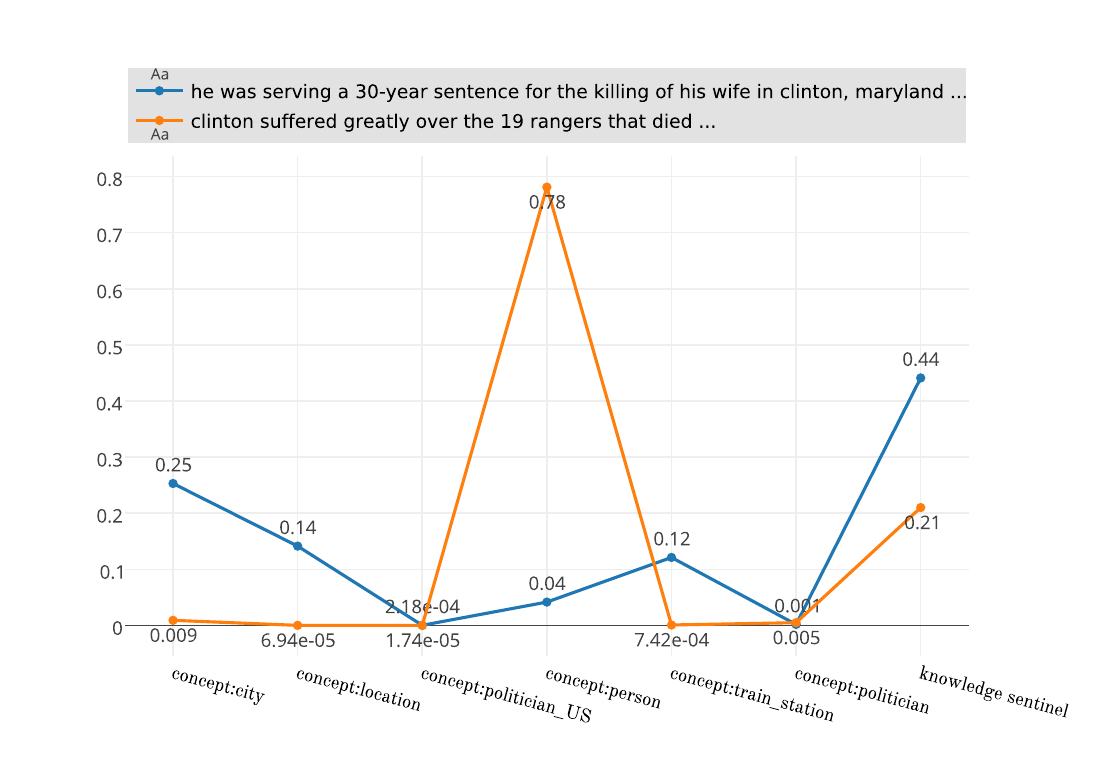}
                \caption{The X-axis represents relevant NELL concepts for the entity mention \textbf{clinton}. The Y-axis represents the concept weights and the knowledge sentinel weight.}
                \label{fig:clinton}
        \end{subfigure}%
         \hfill
        \begin{subfigure}{.45\textwidth}
         \includegraphics[width=\linewidth]{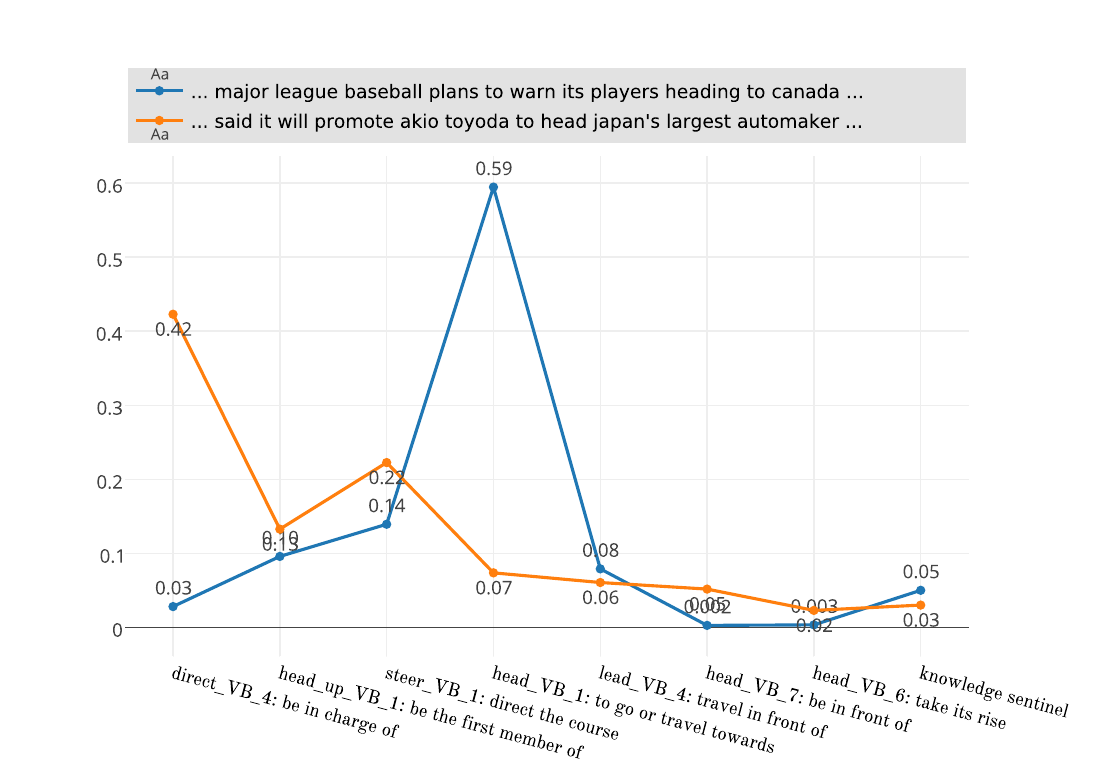}
                \caption{The X-axis represents relevant WordNet concepts for the event trigger \textbf{head}. The Y-axis represents the concept weights and the knowledge sentinel weight. }
                \label{fig:head}
        \end{subfigure}%
\caption{Visualization of the attention weights for KB features learned by KBLSTM-CRF. Higher weights imply higher importance.}
\label{fig:attention}
\end{figure*}

\subsubsection{Results}
We compare our models with the prior state-of-the-art approaches for event extraction, including neural and non-neural ones: {\sc JointBeam} refers to the joint beam search approach with local and global features~\cite{li2013joint}; {\sc JointEntityEvent} refers to the graphical model for joint entity and event extraction~\cite{yang2016joint}; {\sc DMCNN} is the dynamic multi-pooling CNNs in~\newcite{chen2015event}; and {\sc JRNN} is an RNN model with memory introduced by~\newcite{nguyen2016joint}. The first block in Table~\ref{table:event_results} shows the results of the feature-based linear models (taken from~\newcite{yang2016joint}). The second block shows the previously reported results for the neural models. Note that they both make use of gold-standard entity annotations. The third block shows the results of our models. We can see that our KBLSTM models significantly outperform the BiLSTM and BiLSTM-Fea models, which again confirms their effectiveness in leveraging KB information. The KBLSTM-CRF model achieves the best performance and outperforms the previous state-of-the-art methods without having access to any gold-standard entities.

\subsection{Model Analysis}
In order to better understand our model, we visualize the learned attention weights $\boldsymbol{\alpha}$ for KB concepts and the sentinel weight $\beta$ that measures the tradeoff between knowledge and context. Figure~\ref{fig:clinton} visualizes these weights for the entity mention ``clinton". In the first sentence, ``clinton" refers to a {\sc location} while in the second sentence, ``clinton" refers to a {\sc person}. Our model learns to attend to different word senses for 'clinton' (KB concepts associated with 'clinton') in different sentences. Note that the weight on the knowledge sentinel is higher in the first sentence. This is because the local text alone is indicative of the entity type for ``clinton": the word ``in" is more likely to be followed by a location than a person. We find that BiLSTM-Fea-CRF models often make wrong predictions on examples like this due to its inflexibility in modeling knowledge relevance with respect to context. Figure~\ref{fig:head} shows the learned weights for the event trigger word ``head" in two sentences, one expresses a {\sc travel} event while the other expresses a {\sc start-position} event. Again, we find that our model is capable of attending to relevant WordNet synsets and more accurately disambiguate event types.

%% file: sections/conclusion.tex
\section{Conclusion}
In this paper, we introduce the KBLSTM network architecture as one approach to incorporating background KBs for improving recurrent neural networks for machine reading. This architecture employs an adaptive attention mechanism with a sentinel that allows for learning an appropriate tradeoff for blending knowledge from the KBs with information from the currently processed text, as well as selecting among relevant KB concepts for each word (e.g., to select the correct semantic categories for ``clinton" as a town or person in figure~\ref{fig:clinton}). Experimental results show that our model achieves state-of-the-art performance on standard benchmarks for both entity extraction and event extraction.

We see many additional opportunities to integrate background knowledge with training of neural network models for language processing. Though our model is evaluated on entity extraction and event extraction, it can be useful for other machine reading tasks. Our model can also be extended to integrate knowledge from a richer set of KBs in order to capture the diverse variety and depth of background knowledge required for accurate and deep language understanding.
